\documentclass[runningheads]{llncs}
\usepackage{amsmath}
\usepackage[T1]{fontenc}
\usepackage{CJKutf8}
\usepackage{tikz}
\usepackage{calc}
\usepackage{bbding}
\usepackage{txfonts}
\usepackage{multirow}
\usepackage{makecell}

\usepackage{graphicx}
\begin{document}
\title{Co-Driven Recognition of Semantic Consistency via the Fusion of Transformer and HowNet Sememes Knowledge}

\titlerunning{Semantic Consistency Recognition with HowNet}

\author{Fan Chen\inst{1} \and Yan Huang\inst{2}\thanks{Corresponding Author: Yan Huang (Email: platanus@hust.edu.cn)} \and
Xinfang Zhang\inst{1} \and Kang Luo\inst{1} 
\and Jinxuan Zhu\inst{1} \and Ruixian He\inst{1}}

\authorrunning{Fan Chen, Yan Huang, et al.}
%
\institute{School of Mechanical Science and Engineering, Huazhong University of Science and Technology, Wuhan, China \and
School of Artificial Intelligence and Automation, Huazhong University of Science and Technology, Wuhan, China
}

\maketitle              

\begin{abstract}
Semantic consistency recognition aims to detect and judge whether the semantics of two text sentences are consistent with each other. However, the existing methods usually encounter the challenges of synonyms, polysemy and difficulty to understand long text. To solve the above problems, this paper proposes a co-driven semantic consistency recognition method based on the fusion of Transformer and HowNet sememes knowledge. 
Multi-level encoding of internal sentence structures via data-driven is carried out firstly by Transformer, sememes knowledge base HowNet is introduced for knowledge-driven to model the semantic knowledge association among sentence pairs. Then, interactive attention calculation is carried out utilizing soft-attention and fusion the knowledge with sememes matrix. Finally, bidirectional long short-term memory network (BiLSTM) is exploited to encode the conceptual semantic information and infer the semantic consistency. 
Experiments are conducted on two financial text matching datasets (BQ, AFQMC) and a cross-lingual adversarial dataset (PAWSX) for paraphrase identification. Compared with lightweight models including DSSM, MwAN, DRCN, and pre-training models such as ERNIE etc., the proposed model can not only improve the accuracy of semantic consistency recognition effectively (by 2.19\%, 5.57\% and 6.51\% compared with the DSSM, MWAN and DRCN models on the BQ dataset), but also reduce the number of model parameters (to about 16M). In addition, driven by the HowNet sememes knowledge, the proposed method is promising to adapt to scenarios with long text.

\end{abstract}

\section{Introduction}
Semantic consistency recognition (or text semantic matching) task is one of the important tasks of natural language processing (NLP), and can be applied to a wide variety of downstream tasks, such as information retrieval, question answering, dialogue system, machine translation, etc. The input of this task are mainly sentence pairs, different from natural language inference (and text entailment) which aims to recognize the semantic relationship (neutral, entailment, contradiction) between the sentence pairs, the final objective of semantic consistency recognition is to judge whether the semantic meanings of the sentence pairs are consistency or similar. \par

The task of semantic consistency recognition is one of the challenging tasks due to the characteristics of polysemy and synonymy, which are prominent and may lead to ambiguity and misunderstanding especially in Chinese language. 
However, most of the existing algorithms including bag-of-word (BOW), vector space model(VSM), term frequency–inverse document frequency (TF-IDF) and DSSM~\cite{ref_2}, MwAN~\cite{ref_3}, DRCN~\cite{ref_4} cannot catch the semantic meanings accurately, which mainly solve the matching or similarity problem at the lexical level and are difficult to understand text semantics accurately from context. In detail, the existing text matching algorithms based on lexical coincidence and data-driven neural network has the following limitations: \par
\begin{CJK*}{UTF8}{gbsn}
(1) The semantic diversity of word vocabularies. The same word can express different meanings in different context, such as "Apple(苹果)" may denotes a kind of fruit in a menu but can also denotes "Apple Inc(苹果公司)" in the electronic product market.\par

(2) The phrase meaning depends on the order of expression. For a lexical phrase, the meaning expressed will be completely different if exchange the order, such as "奶牛(cow)" and "牛奶(milk)" in Chinese, "one another(彼此,互相)" and "another one(另一个)" in English.\par
\end{CJK*}

In order to solve the above problems, this paper proposes a semantic consistency recognition method based on Transformer~\cite{ref_1} and HowNet~\cite{ref_hownet}, which expands the research on sentence semantic information acquisition. First, we use Transformer to conduct multi-level encoding via data-driven for representation of the internal structure and intrinsic semantics of text sentences. We then introduce an external knowledge base, i.e. HowNet, to conduct knowledge-driven modeling of semantic knowledge association between vocabularies. In addition, soft-attention is exploited to calculate mutual attention and to achieve knowledge fusion with the semantic matrix. Finally, BiLSTM is incorporated to further encode the semantic information of the conceptual level of text and infer the semantic consistency. A number of experiments show that, compared with the existing lightweight models and the pre-training models, the fusion of HowNet sememes knowledge and Transformer's advantage for long text can improve the accuracy of the semantic consistency recognition to a certain extent.\par

The model proposed in this paper has the following innovations:\par
\begin{itemize}
    \item In order to solve the problem of text synonyms, polysemy and difficulty to understand long text for semantic consistency recognition, a co-driven method based on the fusion of Transformer and HowNet sememes knowledge is proposed. In addition to Transformer encoding and model pre-training via data-driven, HowNet sememes are incorporated to enhance the understanding of synonyms and polysemy via knowledge-driven embedding and inference.\par
    \item A technical approach of semantic knowledge fusion is proposed. The multi-level internal structure and semantic information of sentence pairs are encoded through Transformer, and the external knowledge base HowNet is introduced to model the semantic knowledge similarity between sememes sequences. At last, soft-attention is utilized to calculate the interactive attention and conduct knowledge fusion via the sememes matrix.\par
    \item Experiments are conducted on text matching of ant finance, banking finance scenarios and multiple paraphrase identification application. Compared with lightweight models such as DSSM~\cite{ref_2}, MwAN~\cite{ref_3}, DRCN~\cite{ref_4} and pre-training models such as ERNIE, the proposed method can not only improve the accuracy of text semantic consistency recognition effectively, but also reduce the model parameters. It is worth highlighting that our model is capable of adapting to long text. The code will be released at https://github.com/Platanus-hy/sememes\_codriven\_text\_matching.\par
\end{itemize}

\section{Related works}

In recent years, with the rapid development of machine learning, a large number of methods have been proposed to solve the problem of text consistency recognition.
In terms of semantic information acquisition, the classical short text matching model DSSM~\cite{ref_2} solved the problem of dictionary explosion in LSA (latent semantic analysis), LDA (latent dirichlet analysis) and other methods, but also loses context information due to the use of the word bag model. The ESIM~\cite{ref_5} model, proposed in 2016, utilizes the BiLSTM and attention mechanism comprehensively and conducts interaction between sentence pairs in local reasoning for the first time. The DIIN~\cite{ref_6} model proposed in 2018 uses CNN and LSTM for feature extraction, but the author uses both word vectors and local vectors in its input layer, inputs some additional syntactic features, and uses DenseNet for feature extraction. The DRCN model proposed in 2018 learns form DenseNet~\cite{ref_7}'s intensive connection operation for image recognition, it retains the most original information of the text through intensive connection to RNN and continuously adds interactive information to the matrix vector through multiple circulations, and finally outputs via a full connection layer. The KIM~\cite{ref_8} model proposed in 2018 uses the external knowledge base WordNet~\cite{ref_9} to infer the logical relationship between sentence pairs and embed external prior knowledge into the similarity matrix. In the MwAN~\cite{ref_3} model proposed in 2018, the authors use a variety of attention mechanisms (splicing, bilinear, dot multiplication, subtraction) to fully capture the relationship between sentence pairs. Finally, multiple results are weighted and combined to output the final probability through GRU and full connection layers.\par
In terms of sentence structure, CT-LSTM~\cite{ref_10}, which was proposed in 2015, introduces a tree shaped LSTM to solve the problem that LSTM cannot extract the structural information of sentences as well as discussed the long sequence dependency problem. Different from the commonly used RNN sequence modeling, it uses the dependency relationship of sentences as the input of LSTM, which also has some inspiration for future research.\par
With the proposal of BERT~\cite{ref_11} model in 2018, a trend of pre-training models swept the whole NLP world and ranked among the top in the major NLP lists. BERT has a complete Encoder Decoder framework. Its basic composition is Transformer, which is mainly composed of multi-head attention. It is a model built with pure attention, which can solve the problem of long-distance dependency in text sequences, i.e. the attention mechanism is capable to model and remember semantic context over longer distance. The advantage of BERT model is that it can learn more grammatical and semantic information from large corpus, making the output word vector more representative, the larger number of parameters improve the expressive ability and make it perform well in various downstream tasks. In order to accelerate the training speed of the pre-training models and to tackle the high hardware requirements on consumer GPUs, Tim Dettmers~\cite{ref_12} proposed LLM.Int8() for Transformer, which enables ordinary consumer GPUs to use very large models without sacrificing performance. In addition to Transformers, Hanxiao~\cite{ref_13} proposed another simple, attention independent architecture, gMLP, which achieves the same level as Transformers in some indicators of pre-training, and even outperforms Transformers in some downstream tasks.\par

Due to the special polysemy of lexical words, there are certain difficulties in text matching. Different words often express the same meaning, such as "China" and "Huaxia", which are semantically consistent, but not related in terms of grapheme. In order to solve this problem, many researchers choose to use the speech, dependency syntax and other information to calculate the similarity. For example, Yan~\cite{ref_14} and others tried to label the text with part of speech, and only reserved nouns, verbs and adjectives. They obtained word pairs by combining dependency syntax analysis. With PageRank~\cite{ref_15} and degree centrality as indicators, they established a grammar network for a large number of text, and proposed a text similarity calculation method combining syntactic relations and lexical semantics. Yan~\cite{ref_16} proposed a discourse-level text automatic generation model based on topical constraints. The synonymy of the keyword set is used to generate multiple article topic plans. Lin~\cite{ref_17} introduced the concept vector space and represented document as a set of concept words to build a vector space, then calculated the semantic similarity through cosine similarity, which performs better than the bag of word (BOW) model with Word2Vec~\cite{ref_18}.\par

The LET model proposed by Boer~\cite{ref_19} in 2021 utilizes HowNet for text entailment recognition. They transformed the initial vectors of all the sememes for each lexical word with graph attention, then fused the sememes vectors through attention pooling for each word, and obtained the final word vector through the integration of GRU and BERT word vectors. Lexical words are often ambiguous, only a few relevant sememes are essential to identify the correct semantics, which will lead to the fact that the vector of sememes obtained mismatch with the actual sentence, resulting in redundant semantic information. In this paper, the sememes words are filtered in advance before incorporating into the interaction matrix to avoid adding redundant information.\par

\section{Research Method}

This section mainly introduces the co-driven text semantic consistency recognition model based on the fusion of Transformer and HowNet sememes knowledge, and analyzes the model structures and their functions. The proposed model structure is shown in Figure 1, which is divided into 6 layers, namely the Transformer encoding layer, the attention layer, the BiLSTM layer, the pooling layer, the fully connected layer and the prediction layer.

\begin{figure}[ht]
\centering
\includegraphics[width=0.6\textwidth]{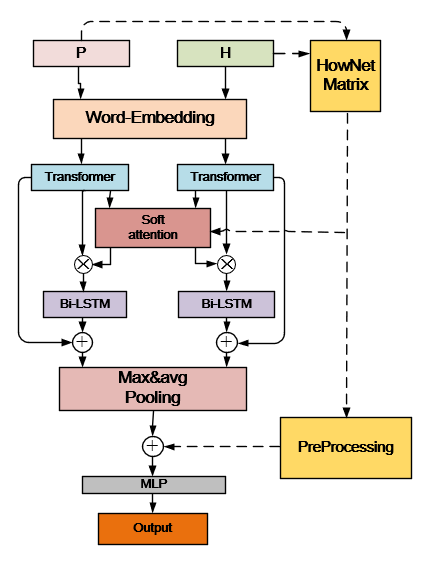}
\caption{Model structure.} \label{fig1}
\end{figure}

\subsection{The Transformer Encoding Layer}
The role of the encoding layer is to model the sequential text to obtain deep semantic information through the neural network. Instead of utilizing the commonly used neural networks such as CNN, LSTM, etc., we exploit Transformer encoder architecture for sentence pair encoding, which is mainly composed of multi-head attention modules and can alleviate the problem of gradient vanishing. The multi-head attention mechanism is formalized as in Eq.(1) to Eq.(4):\par
\begin{equation}
    {Q} = \mathrm{W}^{Q} X
\end{equation}
\begin{equation}
    {K} = \mathrm{W}^{K} X
\end{equation}
\begin{equation}
    {V} = \mathrm{W}^{V} X
\end{equation}

\begin{equation}
    \operatorname{Attention}(Q, K, V)=\operatorname{softmax}\left(\frac{Q K^{T}}{\sqrt{d_{k}}}\right) V
\end{equation}
where $X$ is the input sentence, $W^{Q}, W^{K}, W^{V} \in R^{d_{\text {model}} \times d_{k}}$ are weight matrix, $Q$, $K$, $V$ are embeddings of $X$ for query, key and value respectively, i.e. query for the appropriate key using $Q$ and $K$, then select the corresponding value $V$ via softmax classifier.\par
In addition, the absolute position encodings are calculated as in Eq.(5) and Eq.(6).\par

\begin{equation}
    PE_{(p, 2m)} = \sin \left(\frac{p}{10000^{2 m / d}}\right)
\end{equation}

\begin{equation}
    PE_{(p, 2m+1)} = \cos \left(\frac{p}{10000^{2 m / d}}\right)
\end{equation}
where $d$ denotes the dimension of the word embedding, $p$ denotes the index of the word in the sentence.

\subsection{The Attention Layer}
Attention layer is an important component of text consistency recognition model, which has the advantages of fast, effective and lightweight. Various types of attention mechanisms are proposed in the recent years, such as Soft-attention, Hard attention, Self-attention, etc.
Haili~\cite{ref_20} also proposed an attention mechanism which focus on fine-grained sentiments. This paper adopts the commonly used soft-attention mechanism, but incorporates the semantic matrix generated based on HowNet sememes knowledge, which are balanced by trainable weights $\gamma$ \par
HowNet is a lexical sememes knowledge base for Chinese and English, which discloses the semantic relationships between the concepts and their sememes. This paper mainly uses HowNet to obtain all the lexical sememes corresponding to the words in the sentence pairs. Namely, if two words share the same sememes, the value of their corresponding position in the HowNet sememes matrix will be set to 1, otherwise set to 0. 
\begin{CJK*}{UTF8}{gbsn}
Figure 2 shows an example of sememes interactive calculating for sentence "我们都是中华儿女(We are all childrens of China)" and sentence "华夏起源于先秦(Huaxia originated in the pre-Qin Dynasty)".\par
\end{CJK*}

\begin{figure}
\centering
\includegraphics[width=0.85\textwidth]{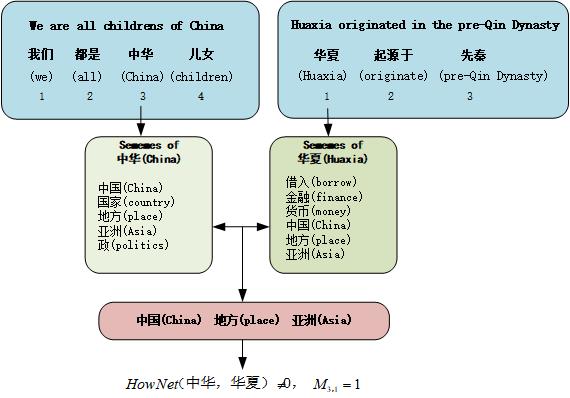}
\caption{Analysis of HowNet semantics.} \label{fig2}
\end{figure}

\begin{CJK*}{UTF8}{gbsn}
The boxes in the top represent the segmented sentences of $P$ and $H$ as input for word embedding, the middle box represents the multiple sememes information corresponding to the each word, and the lower box represents the interaction of the sememes information of two words. 
The HowNet interaction matrix is calculated as in Eq.(7) and Eq.(8). It can be seen from the Figure 2 that the word "China(中华)" has sememes including "China(中国)", “ country(国家)”, “local(地方)” and "Asia(亚洲)","politics(政)", and the word "Huaxia(华夏)" has sememes including "borrow(借入)", "finance(金融)", "money(货币)", "China(中国)", "place(地方)" and "Asia(亚洲)". The intersection of the two words is “China(中国)”, “place(地方)” and “Asia(亚洲)”.\par
\end{CJK*}

\begin{equation}
    M_{i, j}=\left\{\begin{array}{cc}
    1 & \text { HowNet }\left(P_{i}, H_{j}\right) \neq 0 \\
    0 & \operatorname{How} \operatorname{Net}\left(P_{i}, H_{j}\right)=0
\end{array}\right.
\end{equation}
\begin{equation}
    M=\left[\begin{array}{ccc}
    0 & \cdots & 1 \\
    \vdots & \ddots & \vdots \\
    1 & \cdots & 0
    \end{array}\right]
\end{equation}
where $P$ and $H$ represent the embedding results of word segmentation (via tools including Jieba, PKUseg, HanLP, etc.) of the sentence pairs respectively. The vocabularies are vectorized via random initialization when without pre-training. For pre-training, the word embeddings of Bert are used. The generation process of the attention matrix is formalized as Eq.(9): \par
\begin{equation}
    e=PH^{T}+\gamma \cdot M
\end{equation}
where $\gamma$ is a trainable parameter. The attention matrix $e$ not only integrates the structure and semantic information between sentences, but also obtains the synonymous and polysemous relationships of word pairs between sentences. The heat map of matrix change is shown in Figure 4. After injecting the sememes information, the weights of some intersection positions increased, which indicates that the positions reflect the key features of semantic consistency between two sentences. After incorporating the attention matrix, the soft-attention is calculated as follows:\par

\begin{equation}
    \hat{P}=\sum_{j=1}^{l_{h}} \frac{\exp \left(e_{i j}\right)}{\sum_{k=1}^{l_{h}} \exp \left(e_{i k}\right)} P_{t f}, \forall i \in\left[1,2, \cdots, l_{p}\right]
\end{equation}

\begin{equation}
    \hat{H}=\sum_{j=1}^{l_{p}} \frac{\exp \left(e_{i j}\right)}{\sum_{k=1}^{l_{p}} \exp \left(e_{i k}\right)} H_{t f}, \forall i \in\left[1,2, \cdots, l_{h}\right]
\end{equation}

In Eq.(10) and Eq.(11), $P_{t f}$ and $H_{t f}$ are embedding matrix of sentence pairs after encoding via Transformer. $l_{p}$ and $l_{h}$ indicate the length of the sentences as well as the output of the soft-attention module.\par

\begin{figure}
\centering
\includegraphics[width=0.85\textwidth]{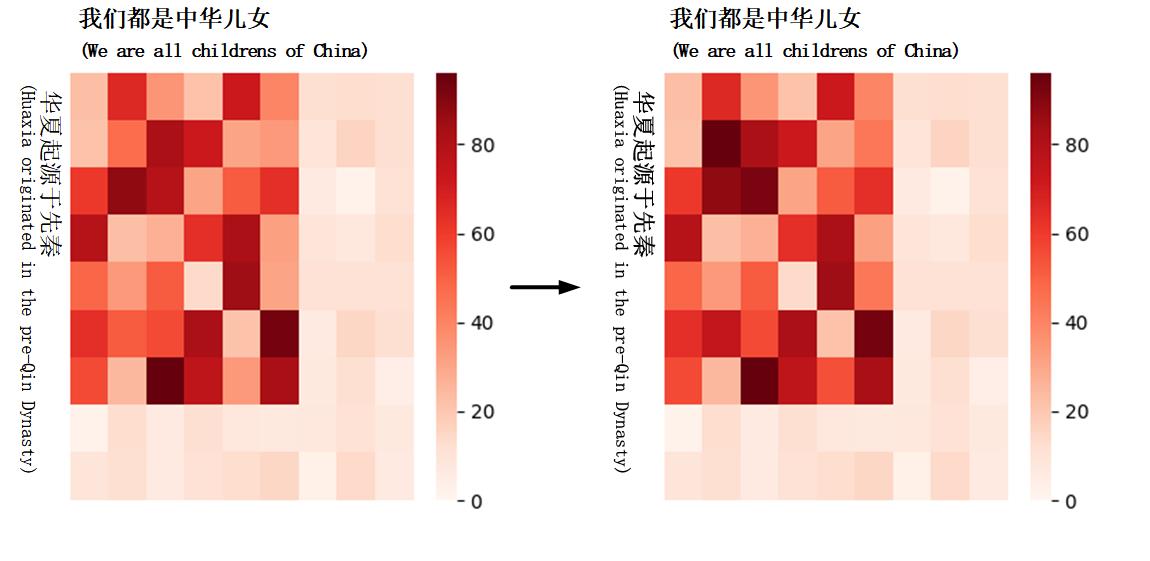}
\caption{Changes of attention matrix.} \label{fig3}
\end{figure}

\subsection{The BiLSTM Layer}
This layer is used to process the output of Transformer layer after the soft-attention mechanism, i.e. $\hat{P}$ and $\hat{H}$. Through the Bidirectional Long Short-Term Memory (BiLSTM), the forward encoding and backward encoding are concatenated to obtain the context information.\par
The output of the Long Short-Term Memory (LSTM) network is as follows:\par

\begin{equation}
    P_{bi-lstm} = \operatorname{BiLSTM}(\hat{P})
\end{equation}

where $\hat{P}$ refers to the output of the sentence $P$ via the soft-attention mechanism, and $P_{bi-lstm}$ refers to the encoding result of $\hat{P}$ via the BiLSTM module. \par

\subsection{Average Pooling and Max-Pooling}
In order to fuse the text information after Transformer and BiLSTM, this model splices multiple inputs through max-pooling and average pooling. The purpose of this layer is to transform the vector dimension of two sentences from $R^{l\times d}$ to $R^{l\times d}$ and facilitate the subsequent input of full connection layer:\par

\begin{equation}
    P_{o}=\left[P_{tf} ; P_{bi-lstm}\right]
\end{equation}
\begin{equation}
    P_{rep}=\left[\operatorname{MaxPool}\left(P_{o}\right) ; \operatorname{AvgPool}\left(P_{o}\right)\right]
\end{equation}
where $P_{tf}$ denotes the output of Transformer encoding module, $P_{bi-lstm}$ denotes the output of BiLSTM module, $P_O$ is the concatenation of $P_{tf}$ and $P_{bi-lstm}$, $P_{rep}$ denotes the result after concatenation of max-pooling and average pooling output of $P_O$.

\subsection{The Fully Connected Layer}
After obtaining the complete sentence vector expression for the sentence pairs, i.e. $P_{rep}$ and $H_{rep}$, the commonly used vector splicing method is to direct splice and input them into the multi-layer feed-forward neural network to obtain the results. When conducting splicing, the information in the HowNet matrix is considered, the sum of two different dimensions in the HowNet matrix is obtained as $HN_{row}$ (sum for rows) and $HN_{col}$ (sum for cols). The final input $H$ of the feed forward neural network is obtained via the concatenation with $P_{rep}$ and $H_{rep}$.\par

\begin{center}
\begin{equation}
\label{eq:Positional Encoding}
\begin{gathered}
HN_{\text {row}}=\operatorname{sum}(M, axis=0) \\
HN_{\text {col}}=\operatorname{sum}(M, axis=1) \\
H=\operatorname{concat}\left(P_{rep};H_{\text {rep}};P_{rep}-H_{rep};HN_{\text {col }};HN_{\text {row }}\right)
\end{gathered}
\end{equation}
\end{center}
where $sum(M, axis)$ denotes summation of $M$ along the axis dimension. For example, $HN_{row}$ denotes the result of the sum of the HowNet matrix along the first dimension, that is, the HowNet information corresponding to sentence $P$. Through vector concatenation, the corresponding semantic information of the two sentences is obtained, in which, $P_{rep} - H_{rep}$ also represents the difference between the two sentence vectors.\par

\subsection{The Prediction Layer}
After obtaining the final sentence vector representation of the sentence pairs, the model uses a two-layer fully connected neural network and a softmax layer to classify the sentence matching results into 1(positive) or 0(negative) as in Eq.(16). The cross entropy loss function is calculated as in Eq.(17).\par
\begin{equation}
    p=softmax(FFN(FFN(H)))
\end{equation}

\begin{equation}
    Loss=\frac{1}{N} \sum_{i}^{N}-\left[y_{i} \times \log \left(p_{i}\right)+\left(1-y_{i}\right) \times \log \left(1-p_{i}\right)\right]
\end{equation}
where $y_{i}$ represents the label of the $i$-th sample $(P_i,H_i)$, the label of positive sample is 1, otherwise 0. $p_{i}$ indicates the probability that the sample is predicted to be a positive sample.\par
In addition to the commonly used cross entropy loss function, we also tried the CoSent loss function, which forces the similarity of positive sample pairs greater than that of negative samples, and makes the distance between positive and negative samples in the vector space as far as possible. The experiment shows that using the CoSent loss function has certain effect on pre-training methods such as BERT and SentenceBERT~\cite{ref_22}, reducing the converge time cost of pre-training models. But for the proposed model in this paper, the effect of CoSent loss reduces compared with cross entropy loss when without pre-training.\par

In the training phase, we used MultiStepLR to dynamically adjust the learning rate, which reduces the learning rate of each parameter group by a decay rate of 0.5 once the number of epoch reaches one of the milestones (i.e. the 20th, 50th, 80th, 100th and 150th iterations of the experiment). By dynamically adjusted the learning rate as the number of iterations increases, the convergence speed of the model increases, and its variation trend is shown in Figure 4.\par

\begin{figure}
\centering
\includegraphics[width=0.6\textwidth]{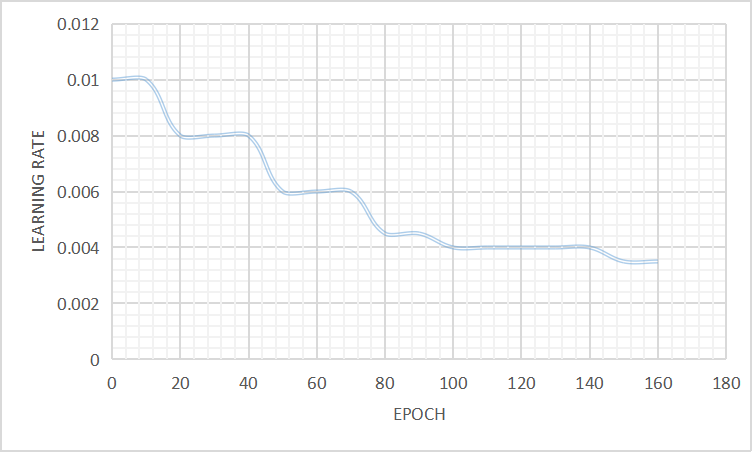}
\caption{The decay trend of learning rate.} \label{fig4}
\end{figure}

\section{Experiments And Analysis}

\subsection{Datasets}

\begin{table}
\centering
\caption{Dataset Size}\label{tab1}
\begin{tabular}{|c|c|c|c|}
\hline
\multicolumn{1}{|l|}{Dataset Name} & \multicolumn{1}{l|}{Training Set Size} & \multicolumn{1}{l|}{Validation Set Size} & \multicolumn{1}{l|}{Test Set Size} \\ \hline
PAWSX                              & 49401                                  & 2000                                     & 2000                               \\
AFQMC                              & 34334                                  & 4316                                     & 3861                               \\
BQ Corpus                          & 100000                                 & 10000                                    & 10000                              \\ \hline
\end{tabular}
\end{table}

In order to verify the effectiveness of the text consistency recognition model based on Transformer and HowNet dual drivers proposed in this paper, this paper conducts experiments on three open datasets respectively. The data sets are respectively PAWSX data set, AFQMC data set and BQ Corpus data set.\par
PAWSX dataset is a multilingual definition pair dataset released by Google. It is characterized by highly overlapping vocabulary, which is helpful to further improve the model's judgment on difficult samples, and can test the model's judgment ability on similar samples. The AFQMC dataset is an ant financial similarity dataset, which contains 34,334 training data, 4316 validation data and 3861 test data. BQ Corpus is the problem matching data set in the banking and financial field, including the problem pairs extracted from the online banking system log of one year. It is the largest problem matching data in the banking field at present, including 10000 training data, 10000 verification data, and 10000 test data.\par

\begin{CJK*}{UTF8}{gbsn}
\begin{table}
\centering
\caption{Example of BQ Dataset}\label{tab2}
\begin{tabular}{|c|c|c|}
\hline
Sentence1 & Sentence2  & label            \\ \hline
\makecell[c]{微信消费算吗 \\ (Does Wechat consumption count)}    & 
\makecell[c]{还有多少钱没还 \\ (How much money is still unpaid)}    & 0(negative) \\
\makecell[c]{下周有什么好产品 \\ (What are the good products next week)} & \makecell[c]{元月份有哪些理财产品 \\ (What financial products are available\\ in January)} & 1(positive)    \\
\makecell[c]{能查账单吗 \\ (May I check the bill)}     & 
\makecell[c]{可以查询账单 \\ (You can check the bill)}     & 0(negative)                \\
\makecell[c]{现在无法借款 \\ (It's unable to borrow now)}     & 
\makecell[c]{QQ有微粒贷平台 \\ (QQ has Weilidai loan product)}     & 0(negative)                \\ \hline
\end{tabular}
\end{table}
\end{CJK*}

\subsection{Experimental Setup}
The experiments of this paper are conducted on a 4-card GPU server with RTX2080ti. The parameters and softwares used for model training are shown in Table 3.\par

\begin{table}
\centering
\caption{Initial training parameters of the model}\label{tab3}
\begin{tabular}{|c|c|}
\hline
Parameter                    & Value   \\ \hline
Word-embedding dim           & 300     \\
Number of hidden layers      & 128     \\
Maximum sequence len         & 100     \\
Batch\_size                  & 64      \\
Number of Transformer layers & 10      \\
Optimizer                    & Adam    \\
Initial learning rate        & 0.01    \\ \hline
Software                     & Version \\ \hline
python                       & 3.6.13  \\
torch                        & 1.10.2  \\
OpenHowNet                   & 2.0     \\
transformers                 & 4.18.0  \\ \hline
\end{tabular}
\end{table}

\subsection{Comparison of experimental results}
In order to verify the actual effect of the model proposed in this paper, three classical text matching models including DSSM, MwAN and DRCN, are selected for comparison without pre-training. For the pre-training model, we chose BERT-wwm-ext, BERT and Baidu ERNIE.\par
The selected data sets are PAWSX, AFQMC and BQ Corpus. In order to ensure the unity of the experiment, all models use the same Jieba vocabulary for the same dataset, and the indicators of comparison are ACC and F1-score.\par

\begin{table}
\centering
\caption{Experimental results on the BQ Dataset}\label{tab4}
\begin{tabular}{|c|c|c|c|}
\hline
Model name           & Pre-trained & Acc              & F1       \\ \hline
DSSM                 & \XSolidBrush           & 77.12            & 76.47    \\
MwAN                 & \XSolidBrush           & 73.99            & 73.29    \\
DRCN                 & \XSolidBrush           & 74.65            & 76.02    \\
Ours                 & \XSolidBrush           & \textbf{78.81}            & \textbf{76.62}    \\ \hline
{Improvement}        & \XSolidBrush        & {+2.19\%}        & +1.96\%  \\ \hline
BERT-wwm-ext         & \CheckmarkBold       & 84.71            & 83.94    \\
BERT                 & \CheckmarkBold         & 84.50            & 84.00    \\
ERNIE                & \CheckmarkBold          & 84.67            & 84.20    \\
Ours-BERT            & \CheckmarkBold           & \textbf{84.82}            & \textbf{84.33}    \\ \hline
{Improvement}        & \CheckmarkBold           & +0.177\%         & +0.464\% \\ \hline
\end{tabular}
\end{table}

\begin{table}
\centering
\caption{Experimental Results on the AFQMC Dataset}\label{tab5}
\begin{tabular}{|c|c|c|c|}
\hline
Model name           & Pre-trained & Acc              & F1       \\ \hline
DSSM                 & \XSolidBrush           & 57.02            & 30.75    \\
MwAN                 & \XSolidBrush           & 65.43            & 28.63    \\
DRCN                 & \XSolidBrush           & 66.05            & 40.60    \\
Ours                 & \XSolidBrush           & \textbf{66.62}            & \textbf{42.93}   \\ \hline
{Improvement}        & \XSolidBrush           & {+0.86\%}        & +5.7\%  \\ \hline
BERT-wwm-ext         & \CheckmarkBold           & 81.76            & 80.62    \\
BERT                 & \CheckmarkBold         & 81.43            & 79.77    \\
ERNIE                & \CheckmarkBold           & 81.54            & 80.81    \\
Ours-BERT            & \CheckmarkBold           & \textbf{81.84}            & \textbf{81.93}    \\ \hline
{Improvement}        & \CheckmarkBold           & +0.097\%         & +1.38\% \\ \hline
\end{tabular}
\end{table}

\begin{table}
\centering
\caption{Experimental Results on the PAWSX Dataset}\label{tab6}
\begin{tabular}{|c|c|c|c|}
\hline
Model name           & Pre-trained & Acc              & F1       \\ \hline
DSSM                 & \XSolidBrush           & 42.64            & 59.43    \\
MwAN                 & \XSolidBrush           & 52.70            & 52.65    \\
DRCN                 & \XSolidBrush           & 61.24            & 56.52    \\
Ours                 & \XSolidBrush           & \textbf{62.55}            & \textbf{59.72}    \\ \hline
{Improvement}        & \XSolidBrush           & {+2.13\%}        & +0.48\%  \\ \hline
BERT-wwm-ext         & \CheckmarkBold           & 77.23            & 76.52    \\
BERT                 & \CheckmarkBold         & 77.06            & 77.16    \\
ERNIE                & \CheckmarkBold          & 78.02            & 77.59    \\
Ours-BERT            & \CheckmarkBold           & \textbf{78.33}            & \textbf{77.96}    \\ \hline
{Improvement}        & \CheckmarkBold           & +0.397\%         & +0.476\% \\ \hline
\end{tabular}
\end{table}

It can be seen from Table 4 that the accuracy of the proposed model in BQ dataset is higher than that of other models. As shown in Table 5, from the perspective of data sets, the results of the three models on AFQMC are not very good. Preliminary analysis shows that the language standardization of the sample data is poor, such as incomplete sentences, such as 
\begin{CJK*}{UTF8}{gbsn}
"可以帮我冻结花呗吗(Can you freeze my Huabei account)" 
\end{CJK*}
and 
\begin{CJK*}{UTF8}{gbsn}
"冻结花呗额度(Freeze the Huabei credit)".
\end{CJK*}
The label is consistent, which leads to poor results in training set and test set. However, the model proposed in this paper integrates Transformer and the HowNet sememes knowledge base, which has a better effect. The results show that for non-standard text, the performance can be improved by obtaining the semantic information of some words and matching them. As shown in Table 6, for the PAWSX dataset where the data samples are difficult to sample, the traditional DSSM model cannot obtain interactive information and context information, so the effect is poor. For hard-negative samples, the highly similar sentence pairs lead to too similar semantic information. The HowNet matrix generated by each pair of sentences is almost identical, so the method of obtaining semantic knowledge is not good for judging the positive and negative samples of difficult samples.\par
From the perspective of error analysis, because we directly used Jieba word segmentation to preprocess the text, the word segmentation error produced by it has different degrees of influence on the experimental results. Although there are segmentation errors, all models use the same vocabulary for the same dataset, and the model proposed in this paper is more effective than others.
\subsection{Ablation Study}
In this section, in order to evaluate the necessity and effectiveness of each module of the model, ablation studies were conducted on the BQ dataset for different structures of the model proposed. Table 7 shows the evaluation results of the impact utilizing different word segmentation tools with or without HowNet. Table 8 shows the evaluation results of the impact with different sequence length. In addition, the number of Transformer layers for text encoding is evaluated in Table 9.\par

\begin{table}
\centering
\caption{Semantic consistency recognition results on the BQ dataset using different tokenizers with or without HowNet.}\label{tab7}
\begin{tabular}{|c|c|c|c|}
\hline
Tokenizer               & HowNet    & Acc             & F1             \\ \hline
\multirow{2}{1.5cm}{Jieba}        & \CheckmarkBold      & \textbf{0.7881} & \textbf{0.7662} \\
                        & \XSolidBrush      & 0.7783          & 0.7624          \\ \hline
\multirow{2}{1.5cm}{PKUseg}       & \CheckmarkBold      & \textbf{0.7869} & \textbf{0.7653} \\
                        & \XSolidBrush      & 0.7792          & 0.7611          \\ \hline
\multirow{2}{1.5cm}{HanLP}        & \CheckmarkBold      & \textbf{0.7853} & \textbf{0.7599} \\
                        & \XSolidBrush      & 0.7735          & 0.7512          \\ \hline
\end{tabular}
\end{table}

It can be seen from the experimental results in Table 7 that using HowNet can improve the performance of the model to some extent. Compared with not using HowNet, the accuracy of various word segmentation tools is improved. If there are some words with complex semantics in the data sample, the introduction of external knowledge base can significantly improve the sensitivity of the model to polysemy and synonym, and can significantly improve the performance of the model.\par

\begin{table}
\centering
\caption{Semantic consistency recognition results on the BQ dataset for different sequence length with or without HowNet.}\label{tab8}
\begin{tabular}{|c|c|c|c|}
\hline
Sequence length & HowNet & Acc              & F1     \\ \hline
1$\sim$15       & \CheckmarkBold      & \textbf{0.7869} &\textbf{0.7662}  \\
                & \XSolidBrush      & 0.7763          & 0.7521 \\ \hline
15$\sim$50      & \CheckmarkBold      & \textbf{0.7884} & \textbf{0.7684} \\
                & \XSolidBrush      & 0.7792          & 0.7545 \\ \hline
\end{tabular}
\end{table}

According to the results in Table 8, HowNet can effectively improve the performance of both text with a length of less than 15 and text with a length of more than 15 and less than 50, and can obtain more valid semantic information in longer text to achieve better results. After the longest text segment experiment in the dataset, the longest text length that this model can handle is 50 on the basis of ensuring the experimental effect.\par

\begin{table}
\centering
\caption{Semantic consistency recognition results on the BQ dataset with different number of Transformer encoding layers.}\label{tab9}
\begin{tabular}{|c|c|c|}
\hline
Number of Transformer layers & Acc     & F1     \\ \hline
2                            & 0.7433 & 0.7353 \\
4                            & 0.7648 & 0.7572 \\
6                            & 0.7752 & 0.7731 \\
8                            & 0.7853 & 0.7658 \\
10                           & 0.7881 & 0.7662 \\ \hline
\end{tabular}
\end{table}

From the result data in Table 9, the higher the number of Transformer layers, the better the effect of the model. By stacking the Transformer coding layer, the performance of the model can be improved to a certain extent, but at the same time, the number of parameters of the model and the training time of the model will be significantly increased, and the convergence speed will also be significantly slower. We take the model with 6 coding layers as the optimal model, and its parameter quantity is 16M, while the parameter of the DRCN model which performs the best in the models without pre-training is 19M.
During training, the changes of trainable parameters are as follows:/par

\begin{figure}
\centering
\includegraphics[width=0.75\textwidth]{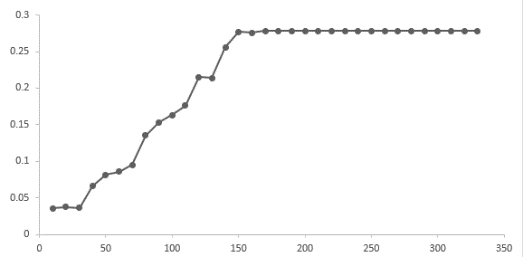}
\caption{Variation diagram of trainable parameters} \label{fig5}
\end{figure}

It can be seen from Figure 5 that during the experiment, by observing the changes of the trainable parameters of the attention matrix, the weight of the sememes information matrix obtained by HowNet in the attention matrix is gradually increased with the increase of the number of iterations, which indicates that the HowNet sememes information generated by the original text has a positive role in improving the model effect.

\section{Conclusion}
This paper proposed a new text semantic consistency recognition model based on Transformer and HowNet, which uses the HowNet sememes knowledge to tackle the synonyms and polysemy problems for semantic matching of sentence pairs. The experiments on datasets including BQ, AFQMC and PAWSX show that, compared with models without pre-training including DSSM, MwAN and DRCN, as well as pre-training models such as ERNIE, the proposed method has a certain improvement. By stacking Transformer layers, the performance of semantic consistency recognition can be effectively improved while introducing more parameters to some extent. In contrast, the model proposed in this paper has fewer parameters and better performance. Compared with LET, which also uses HowNet as the external information base, this paper filters the irrelevant semantic information while utilizing HowNet sememes information, avoiding the impact of redundant sememes on the results, which is more accurate and more intuitive. Experiments show that the model proposed in this paper can improve the accuracy of consistency recognition effectively using HowNet sememes knowledge, and can also adapt to long text scenarios within 50 characters. There are obvious improvements either with the lightweight model or with the pre-training model.\par
In the future, we will study the knowledge limitation of language expression in depth, i.e. a sentence is wrong in combination with common sense, but is correct in terms of morphology and syntax. For example, "the sun travels around the earth" et al.
Thus, more work is needed to supplement the external knowledge, we will continue to utilize commonsense knowledge to further improve the semantic understanding of raw text.

\par

%
%
%
%

\end{document}